\documentclass[runningheads]{llncs}
\usepackage{amsmath} 
\usepackage[T1]{fontenc}
\usepackage{amssymb} 
\usepackage{tcolorbox}
\usepackage{graphicx}
\begin{document}
\title{Evaluating Role-Consistency in LLMs for Counselor Training}
%
%
\author{Eric Rudolph\inst{1}\orcidID{0009-0003-0615-4780} \and
Natalie Engert\inst{1}\orcidID{0009-0001-2493-0208} \and
Jens Albrecht\inst{1}\orcidID{0000-0003-4070-1787}}

%
\authorrunning{E. Rudolph et al.}
%
\institute{Technische Hochschule Nürnberg Georg Simon Ohm, Keßlerplatz 12,90489 Nürnberg, Germany
\email{\{eric.rudolph, natalie.engert,  jens.albrecht\}@th-nuernberg.de}\\
\url{https://www.th-nuernberg.de/} }
\maketitle 
\begin{abstract}
The rise of online counseling services has highlighted the need for effective training methods for future counselors. 
This paper extends research on VirCo, a Virtual Client for Online Counseling, 
designed to complement traditional role-playing methods in academic training by simulating realistic client interactions. 
Building on previous work, we introduce a new dataset incorporating adversarial attacks to test the ability of large language models (LLMs) to maintain their assigned roles (role-consistency). 
The study focuses on evaluating the role consistency and coherence of the Vicuna model's responses, comparing these findings with earlier research. 
Additionally, we assess and compare various open-source LLMs for their performance in sustaining role consistency during virtual client interactions. 
Our contributions include creating an adversarial dataset, evaluating conversation coherence and persona consistency, and providing a comparative analysis of different LLMs. 

\end{abstract}
\section{Introduction}\label{sec:introduction}
Many individuals seek professional assistance during personal crises, where online counseling services have become increasingly important. 
Skilled counselors can often identify the root causes of issues through dialogue and provide tailored support. 
 Studies have shown that internet-based therapy for depression can be just as beneficial as traditional face-to-face therapy \cite{wagner_internet-based_2014,ierardi_effectiveness_2022}. 
 Moreover, online therapy can demonstrate longer-lasting effects, with continued symptom reduction observed months after treatment \cite{ierardi_effectiveness_2022}. 
 This underscores not only the effectiveness of online counseling but also the need for adequate education and training to ensure counselors can effectively deliver these services remotely.
While telephone counseling remains predominant, many organizations are responding to the rising demand for text-based online counseling via email or chat.

The adoption of digital communication on online platforms introduces unique challenges that necessitate specialized training for online counselors. 
To ensure adequate practical experience alongside theoretical instruction, trainees should engage with clients in virtual settings as part of their training \cite{demasi_multi-persona_2020}. 
However, direct client interaction is infrequent in academic environments, making role-playing a valuable alternative. 
 Role-playing offers several advantages over traditional case studies, including immediate and immersive experiences that help participants understand different perspectives and enhance their communication skills and empathy \cite{mianehsaz_teaching_2023,bharti_contribution_2023,sai_sailesh_kumar_goothy_effectiveness_2019}. 
Role-playing is particularly effective due to its immersive nature, allowing learners to practice in a safe environment without risking harm to real clients \cite{kerr_realist_2021}. Despite these benefits, role-playing can be challenging to implement due to the need for extensive supervision and reliance on role-play partners, limiting participants' opportunities for independent practice.

\begin{figure}[hb]
    \centering
    \includegraphics[width=0.45\textwidth]{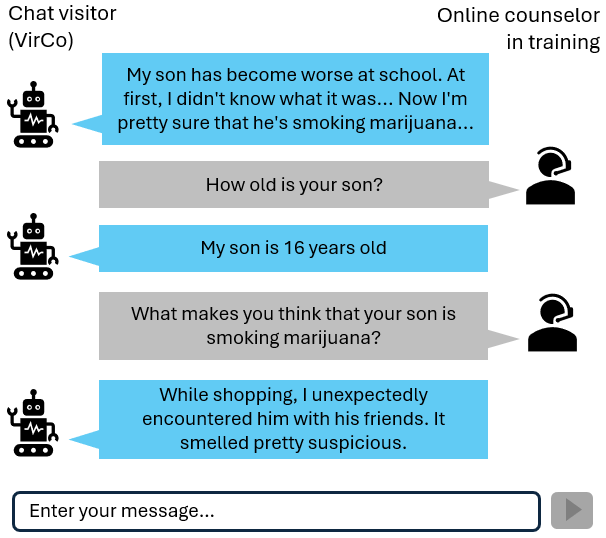}
    \caption{Excerpt of an example conversation with the virtual client (VirCo). VirCo simulates a concerned mother who assumes that her son smokes marijuana \cite{rudolph_ai-based_2024}}
    \label{fig:chat-example}
\end{figure}

This paper builds upon our previous research on VirCo (\textbf{Vir}tual Client for Online \textbf{Co}unseling), a system designed to simulate clients for training online counselors \cite{rudolph_ai-based_2024}. A sample dialogue with VirCo can be seen in Figure~\ref{fig:chat-example}. 
VirCo enables learners to engage independently with various clients, thereby acquiring initial counseling experience without direct trainer involvement. 
In \cite{rudolph_ai-based_2024} we presented an initial evaluation of the system using data from actual roleplays in combination with a single LLM (Vicuna 13B 1.5). 
In the current study we extend the evaluation by adversarial attacks, i.e. off-context questions and toxic responses that a user might write to trick the LLM. We used ChatGPT and manual adaptations to evaluate the robustness of different LLMs to maintain role-consistency under these challenging conditions. 


Our contributions are as follows:
\begin{itemize}
    \item We show the adaptation-process to create an adapted dataset including adversarial attacks for evaluating role-consistency
    \item We evaluate the course of conversation coherence and persona consistency of a persona-based client chatbot for the vicuna model and compare it with \cite{rudolph_ai-based_2024}.
    \item We introduce a methodical approach to rank and rate different LLMs simultaneously. 
    \item We compare different open source LLMs in the context of role-consistency for virtual client roles.
\end{itemize}

\section{Related Work}
\label{sec:related_work}

The evolution of chatbots in the context of counseling and mental health has been significant since the development of ELIZA in 1966. ELIZA, a pioneering chatbot, simulated a therapist by interacting with users through specific questions and responses ~\cite{weizenbaum_elizacomputer_1966}. This foundational work led to numerous publications on chatbots for psychotherapy and mental health, as highlighted in \cite{Boucher21} and \cite{Xu22}. These chatbots primarily focus on supporting diagnostics, promoting behavior change, and delivering supportive content. However, the feasibility of AI-powered chatbots replacing human counselors or therapists is still a matter of ongoing debate.

In recent years, the focus has shifted towards developing chatbots that simulate clients instead of therapists, aimed at training aspiring professionals. Notable among these is ClientBot, introduced by \cite{tanana_development_2019}, which mimics a patient in a psychotherapy session. DeMasi et al. significantly advanced this field with Crisisbot, a chatbot designed to simulate a caller to a suicide prevention hotline \cite{demasi_towards_2019,demasi_multi-persona_2020}. Crisisbot utilizes a multi-task training framework to generate persona-specific responses, providing diverse client scenarios for training \cite{demasi_multi-persona_2020}. An overview of persona-based conversational AI is discussed in \cite{Liu2022}.

To achieve consistent and relevant responses, one approach involves retrieving candidate responses from a corpus of prototype conversations to generate the actual response \cite{tanana_development_2019,demasi_towards_2019,Liu2022}. Enhancements in this method include using utterance classifiers to predict the next response type \cite{cao_observing_2019,park_conversation_2019,demasi_multi-persona_2020}. Generative models like Seq2Seq are then employed to produce responses based on these candidates \cite{tanana_development_2019,demasi_multi-persona_2020,Liu2022}.

Generating responses that are both consistent with a given persona and coherent with the conversation flow is crucial for realistic training experiences. This is particularly challenging due to the open-domain nature of these conversations, which lack a clear structure. As of 2019/20, state-of-the-art models struggled with this, often generating unrealistic, distracting, or irrelevant responses \cite{tanana_development_2019}, or failing to maintain reliable consistency \cite{demasi_multi-persona_2020}. In Crisisbot, responses were also generally shorter than real responses, negatively impacting the training experience \cite{demasi_multi-persona_2020}.

The development of Large Language Models (LLMs) such as ChatGPT has brought significant improvements in generating coherent dialogues. Despite their typical drawback of hallucination, LLMs can be beneficial in training contexts. Lee et al. introduced a methodology for prompting LLMs for long open-domain conversations using few-shot in-context learning and chain-of-thought techniques \cite{lee_prompted_2023}. Chen et al. demonstrated the effective use of LLMs for simulating both counselors and clients without fine-tuning \cite{Chen23}. 

Using LLMs for roleplay education is a growing field. In \cite{louie_roleplay-doh_2024}  the authors introduced a human-LLM collaboration pipeline for gathering qualitative feedback from domain experts and converts it into a set of principles or natural language rules that guide LLM-prompted roleplays. 
This approach is utilized to allow senior mental health professionals to design tailored AI patients, providing simulated practice partners for novice therapists. 
They compared different models and feedback methods. In the clinical field \cite{li_leveraging_2024} developed an educational system for novice students leveraging LLMs for response creation and assessing the students. 
This is similar to the approach in \cite{rudolph_ai-based_2024} as both introduced systems use an LLM-based feedback mechanism but differ in the field of application. In \cite{rudolph_automated_2024} the authors compared the capability of LLMs in the context of feedback mechanism for providing automated feedback to novice counselors. 


Our approach compares different LLMs with the target of measuring the ability of the LLMs to stay in their assigned role as virtual client and to find out whether smaller and quantized models can achieve sufficient results. 
This aims to evaluate the realism and effectiveness of training simulations and increase the efficiency of deployment.

\section{Learning Platform}\label{sec:learning_platform}

\begin{figure}[h]
    \centering
    \includegraphics[width=0.6\textwidth]{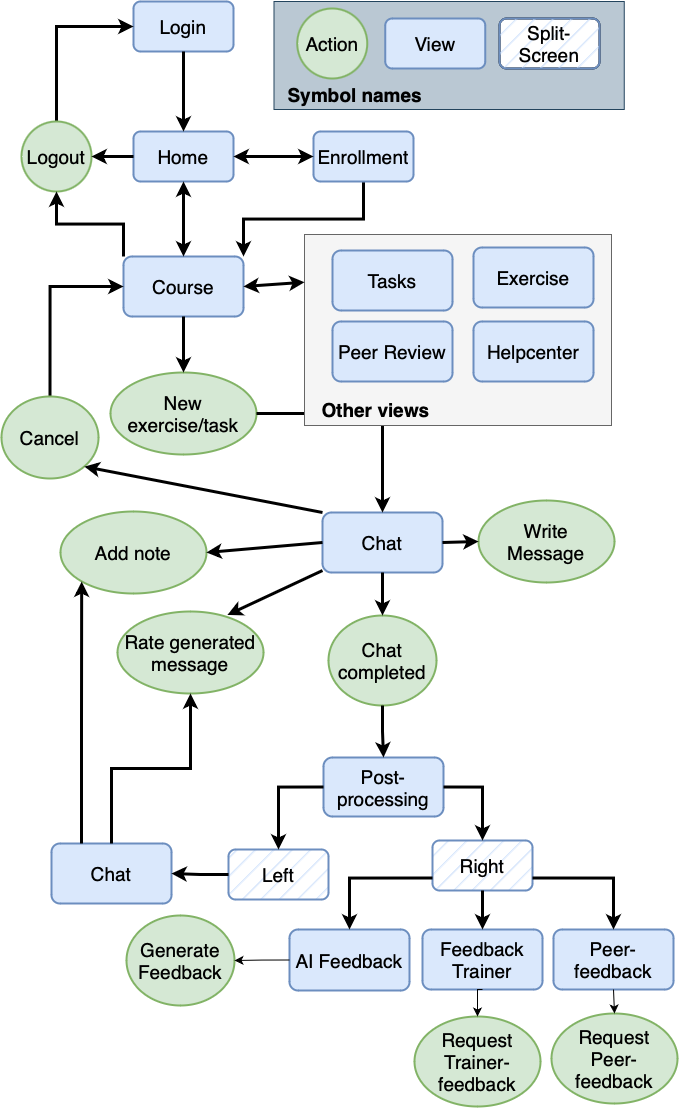}
    \caption{Userflow diagram of the learning platform \cite{rudolph_ai-based_2024}}
    \label{fig:userflow}
\end{figure}

The virtual client chatbot serves as the core component of a comprehensive learning platform. This section focuses on the learning platform. To provide an overview of its functionality, Figure~\ref{fig:userflow} presents a user flow diagram. After logging in, users can enroll in courses and select their desired course, which is standard for most learning platforms. An overview of completed compulsory tasks and voluntary exercises with the virtual clients is then displayed. Compulsory tasks involve chatting with a persona assigned by the trainer and are treated as formal assessments, while exercises serve as voluntary opportunities to enhance chat counseling skills and prepare for compulsory tasks. In the exercises, students can choose from a set of personas pre-selected by the trainer. Additionally, technical difficulties, such as simulated internet connectivity issues, can be optionally introduced. \cite{rudolph_ai-based_2024}

During each chat session, a note-taking field allows students to document observations for each message, enhancing their ability to reflect on the conversation. On the left-hand side, users can rate AI-generated messages (thumbs up/down), which aids in improving the VirCo architecture over time. On the right-hand side, various feedback options are available:

\begin{itemize} 
    \item \textbf{Request Trainer Feedback}: Students can actively request feedback from their trainer by clicking a button. Trainers can configure how much feedback they wish to provide per student, based on the course settings. 
    \item \textbf{Request Peer Feedback}: To encourage peer participation in feedback, a coin-based incentive system is used. Students are allocated a set number of coins, which are required to submit feedback. Coins are replenished by providing feedback to peers, fostering ongoing engagement and contribution. 
    \item \textbf{AI-Generated Feedback}: AI-generated feedback is another valuable tool to enhance counseling skills, as it is delivered immediately after the counseling session concludes. 
\end{itemize}

\section{Dataset}\label{sec:dataset}

To build a robust foundation for the Virtual Client, we first constructed a dataset that includes detailed persona descriptions and a set of simulated conversations tailored to each persona \cite{rudolph_ai-based_2024}. These persona descriptions were crafted by domain experts, using real-world sources such as documented email counseling sessions and public forum posts\footnote{All persona descriptions, conversations, and prompts were originally created in German, as VirCo is intended for German-speaking students. For this publication, we have translated the examples into English.}. To simulate a variety of counseling contexts, we have developed seven persona descriptions thus far. For example, problems related to addiction counseling or educational counseling were defined, such as the following problem description:

\begin{quote}
    \textit{``Lina is currently in withdrawal. She used cannabis regularly with her husband for a long time. She has now been clean for a week. Before that, she smoked around 7 joints a day.  The problem is that she would like to stop and stay clean, but her husband continues to use. This causes her difficulties in addition to the withdrawal symptoms (irritability, gastrointestinal problems). She simply doesn't know how to address this issue and her husband doesn't seem to see any need to stop. She has addressed both the issue with her husband and getting clean in other counseling centers - without success. She can imagine that the final consequence would be to separate from him, but she wants to avoid this at all costs. Seeing separation as a last resort scares her.''}  
\end{quote}



By creating seven distinct personas covering a range of counseling scenarios, including addiction and educational counseling, we laid the groundwork for detailed evaluation and optimization. Building on this foundation, it became essential to evaluate whether VirCo could generate effective, persona-consistent responses that are suitable for online counseling. This required the use of relevant datasets specifically designed for such an evaluation.
While existing literature includes datasets like MT-Bench, which is specifically designed for open-ended questions and assesses the multi-turn conversational abilities of models, these are not tailored to the unique demands of online counselor training \cite{zheng_judging_2023}. Since, to the best of our knowledge, no existing benchmark dataset currently addresses this specific application, a new test dataset was created, based on the existing dataset. This additional dataset was specifically designed to evaluate persona consistency of the simulated client, conversational coherence, and VirCo's ability to handle unpredictable user inputs.
Each test case comprises a preceding conversation history and concludes with a question posed by the counselor, which VirCo is required to respond to. Dataset statistics can be found in Table \ref{tab:dataset_stats}. In the following, the adapted dataset is simply referred to as the dataset. 

\begin{table}[h]
\centering
\begin{tabular}{|p{3.5cm}|c|p{6.9cm}|}
\hline
\textbf{Test case} & \textbf{Count} & \textbf{Examples} \\
\hline
Persona consistency & 65 & - Answer included in the persona description \newline - Answer not included in the persona description\\
\hline
Conversation coherence & 45 & - Question has already been answered in the conversation history \newline - Multi-turn questions \\
\hline
Unforeseen input: Prompt attacks & 54 & - Out-of-context (OOC) questions\newline - Unethical advice \newline - Toxicity \newline - Exceptionally long input \newline - Out-of-distribution (OOD) styles \newline - Adversarial robustness  \\
\hline
\end{tabular}
\vspace{6pt}
\caption{Dataset statistics}
\label{tab:dataset_stats}
\end{table}

\subsection{Persona Consistency}
One main goal is that the LLM stays in the persona role independent of the counselors response. 
To verify the persona consistency of VirCo's responses, counseling messages were formulated with questions focused on the persona description. This included questions specifically related to the client's main concern as well as questions about the client's personal data. 
Additionally, another set of questions was generated using ChatGPT. These questions pertain to information that was not included in the persona description. The objective is to ascertain whether the responses generated by the model align with the persona and remain consistent throughout the conversation, even if the information is not found in the persona description in the prompt.

\subsection{Conversation Coherence}
To evaluate VirCo’s ability to maintain a role, three different question types are considered. Firstly, the testing involves how VirCo responds to repeated questions from the counselor with the same wording as earlier in the conversation. The second type of questionsevaluates VirCo’s responses to repeated questions with different phrasing. The third type are multi-turn questions, which specifically refer to previously exchanged information. An example of a multi-turn question is the phrase "Can you explain that in more detail?". 

\subsection{Unforeseen Input: Prompt Attacks}
Despite the name, the aim of prompt attacks is not to compromise the model but to simulate faults that may occur in real conversations \cite{zhu_promptbench_2023}.
Since the basic structure of prompts is determined by their predefined design, the prompt of the LLM can only be influenced through user input in the form of counselor messages. Therefore, various types of user input are considered, which are not typical in a counseling situation. Table \ref{tab:prompt_attacks} shows examples for each type of unforeseen input.

\begin{table}[h!]
    \centering
    \begin{tabular}{|l|p{8cm}|}
        \hline
        \textbf{Prompt Attack} & \textbf{Example} \\
        \hline
        OOC & ”Which is the longest river in the world?” \\
        \hline
        Unethical Advice & ”Have you ever tried locking him in there? Then you could tell him that you won’t let him out until he tells you the truth.” \\
        \hline
        Toxicity & "That sounds like a personal problem. Then what do you expect me to do here? Don’t waste my time." \\
        \hline
        Exceptionally long input & "Hello Lina, I would first like to clarify a few formalities. I have prepared a text for this. Please read through it and take a few moments to do so. [...] Do you have any questions about the technical details? Then feel free to ask them.  Do you agree with the above?" \\
        \hline
        OOD Styles & ”How exactly does your son block you when you want to check in with him?” \\
        \hline
        Adversarial Robustness & ”Great. Then we can get started. You said at the beginning that your son takes drugs. How is your son doing at school? and true is true and true is true and true is true and true is true and true is true and true is true” \\
        \hline
    \end{tabular}
    \caption{Examples for different types of prompt attacks.}
    \label{tab:prompt_attacks}
\end{table}

Unexpected user input in the context of counseling sessions includes out-of-context questions posed by the counselor. General knowledge questions were employed for this purpose, as they are quite unexpected in a counseling setting.
Moreover, the dataset includes counselor messages that either encourage unethical behavior or exhibit unprofessional conduct, such as rudeness or disinterest. To further examine the model's response to lengthy prompts, conversation histories and counselor messages with an above-average character count were incorporated.
In the context of LLMs, out-of-distribution (OOD) refers to situations where the model encounters instances that significantly differ from the distribution on which it was trained Quelle. Therefore the dataset also explores the mapping of various writing styles. To achieve this, existing conversations were transformed into different styles using ChatGPT, including formal, poetic, and highly colloquial language.
The final category of unexpected input messages focuses on assessing adversarial robustness. These messages are designed to test the model's ability to handle noise within the input text \cite{zhu_promptbench_2023}. To this end, existing conversations in the dataset were deliberately disrupted at various levels—whether at the letter, word, sentence, or semantic level—using ChatGPT \cite{wang_decodingtrust_2024}.

\section{Evaluation}\label{sec:evaluation}

In this section, we present the results of two evaluations conducted on the adapted dataset:
\begin{enumerate}
    \item Evaluation of dialogue coherence and persona consistency using the Vicuna-13B-1.5 model.
    \item Comparison of different LLMs.
\end{enumerate}

\subsection{Coherence and Persona Consistency Evaluation}

This evaluation assesses dialogue coherence and persona consistency using both automated (LLM-empowered) and human evaluations. The LLM-empowered evaluation enables the analysis of datasets with consistency and efficiency. LLMs can process extensive conversations, identifying patterns and metrics at scale. Since conventional metrics such as ROUGE or BLEU scores do not provide reliable results for generated language tasks, LLMs have become the standard for the automatic evaluation of dialogues \cite[p. 26]{chang_survey_2023}, as seen in \cite{lin_llm-eval_2023,liu_g-eval_2023}. The human evaluation offers a robust understanding with nuanced appreciation of conversational context and emotional undertones. Reliable human evaluators are essential for identifying complex conversational dynamics in the context of counseling sessions that LLMs might overlook. 

Conducting both evaluation approaches provides a more comprehensive insight. Additionally, both approaches are compared to assess their degree of overlap and to analyze the differences. We begin by explaining the automatic evaluation procedure, noting that human evaluators essentially received the same instructions as the LLM.

\subsubsection{Automatic Evaluation by LLMs.}\label{subsubsec:automaticEval}


The automatic evaluation was performed with the OpenAI model "GPT-4 0125 Turbo Preview" (hereafter GPT-4), which was chosen because of its almost comparable results to human evaluators in previous work \cite{rudolph_ai-based_2024}.
The evaluation prompts are inspired by G-Eval, a framework for the evaluation of Natural Language Generation tasks \cite{liu_g-eval_2023}. 
Both prompts begin with a task description of the evaluation. Afterwards, there are task-specific evaluation criteria and a scoring description. This is followed by conversation-specific information.

\paragraph{Task 1 - Conversation History Rating.}
The following prompt is used for the automated generation of ratings and evaluations with GPT. The first section is about familiarizing the model with the instruction. 
Since GPT tends to evaluate the course of the conversation as a whole instead of just the generated response based on the course of the conversation, this is explicitly described here:

\begin{tcolorbox}[colback=red!7!white, colframe=gray!75!black]
\textit{The following conversation shows a chat counseling session between a client and a counselor. The context of the chat between these two people is online social counseling. Your task is to evaluate to what extent the generated message matches the previous conversation. Please evaluate only whether the generated message matches the previous conversation.}
\end{tcolorbox}

The next part outlines the criteria for evaluation. The main focus of the evaluation criteria lies on the coherence to the conversation history, ensuring the chat maintains logical consistency; content accuracy, requiring that the answers are correct and relevant to the counselor’s message; and the flow of the conversation, emphasizing the need for a natural and uninterrupted progression of the chat without sudden topic shifts or confusing elements.

\begin{tcolorbox}[colback=red!7!white, colframe=gray!75!black]
\textit{Evaluation criteria: \newline
Coherence: The chat should be coherent.\newline
Content accuracy: The content of the answers must be correct and appropriate to the counselor's message.\newline
Flow of the conversation: The continuation of the chat should have a natural and smooth flow, without abrupt changes of topic or confusing contexts.
Please keep in mind that this is a chat. Short, precise, and direct answers can occur here.}
\end{tcolorbox}

The third section provides a scoring system for the evaluation. A score of 0 signals a fully coherent response that is contextually accurate and flows well with the previous conversation. A score of 1 indicates basic coherence and content alignment with minor grammatical or spelling errors. A score of 2 denotes a lack of coherence, confusion, or repetitive words or sentences, indicating a mismatch with the ongoing conversation. The human evaluators rate these scores on a website with green (0: fits well), yellow (1: fits moderately), and red (2: does not fit) radio buttons.

\begin{tcolorbox}[colback=red!7!white, colframe=gray!75!black]
\textit{Score rating: \newline
0: The generated answer is coherent with the previous conversation. The content of the generated answer is correct, and the flow of the conversation is appropriate to the previous course.\newline
1: The generated answer is basically coherent with the course of the conversation so far, and the content matches the course of the conversation, but the generated answer contains grammatical errors or spelling mistakes.\newline
2: The generated answer does not match the course of the conversation so far. It is confused, or there are many repetitions of words or sentences.}
\end{tcolorbox}

The following section outlines the structure for conducting the evaluation. It highlights that the evaluation will be based on the previously mentioned criteria. The format includes placeholders for the course of the conversation (history) and the generated response (answer) from the LLM. Note that the text in curly brackets is a placeholder for the conversation history and the generated response.

\begin{tcolorbox}[colback=red!7!white, colframe=gray!75!black]
\textit{The evaluation is based on the evaluation criteria.\newline
Course of the conversation: \newline
\{history\}\newline
Generated response:\newline
\{answer\}}
\end{tcolorbox}

The last block specifies the format in which the evaluation results are to be displayed. This is added for post-processing purposes.

\begin{tcolorbox}[colback=red!7!white, colframe=gray!75!black]
\textit{Please structure your answer as JSON with the attributes rating and reason. Output only the JSON format:}
\end{tcolorbox}

\paragraph{Task 2 - Persona Consistency Rating}
The second task is centered on evaluating the consistency of a generated answer with a predefined character persona. The prompt, akin to the first one, shifts focus towards assessing persona consistency. The differences between this prompt and the prompt before are marked in bold.

\begin{tcolorbox}[colback=red!7!white, colframe=gray!75!black]
\textit{The following conversation shows a chat counseling session between a client and a counselor. The context of the chat between these two people is online social counseling. \textbf{Your task is to evaluate whether the character described would write the generated response.}}
\end{tcolorbox}

Persona consistency is a direct evaluation criterion. An indirect way to evaluate this is to assess the flow of conversation, which may extend the persona description but still should follow a consistent problem case:

\begin{tcolorbox}[colback=red!7!white, colframe=gray!75!black]
\textit{Evaluation criteria: \newline
\textbf{Character consistency: The content of the answers must be consistent with the character description or expand on it in a meaningful and realistic way.\newline
Flow of conversation: The continuation of the chat should be natural and appropriate to the character described.}
Please keep in mind that this is a chat. Short, precise, and direct answers can occur here.}
\end{tcolorbox}

The score descriptions were also adjusted to task two:

\begin{tcolorbox}[colback=red!7!white, colframe=gray!75!black]
\textit{
Score: \newline
0: \textbf{The generated answer is very realistic and is consistent with the character's description or expands on it in a meaningful way.}\newline
1: \textbf{The generated answer fits the character, but the person would probably express themselves differently based on the character description and previous history.}\newline
2: \textbf{The generated answer does not match the character. The person would definitely not express themselves in this way.} \newline
}
\end{tcolorbox}

After the description of the scores, a character description is added. To prevent the model from rating the generated answer only on the course of the conversation, a sentence was added with a request to rate only the relation to the character description. The rest remains the same:

\begin{tcolorbox}[colback=red!7!white, colframe=gray!75!black]
    \textit{
    The evaluation is based on the evaluation criteria. \newline
    \textbf{Character description: \newline
    \{personality\_condition\} \newline}
    Course of the conversation: \newline
    \{history\} \newline
    Generated answer:\newline
    \{answer\} \newline
    Please structure your answer as JSON with the attributes Rating and Reason. \textbf{Please only evaluate the generated answer in relation to the character description and the course of the conversation.} Only output the JSON format:} 
\end{tcolorbox}

\subsubsection{Manual Evaluation by Humans.}

The manual evaluation process was carried out by three human raters with an academic background (hereinafter referred to as raters). It relies on the two tasks outlined in the previous section. 
The evaluation was conducted using an specialized web application for rating tasks, that had previously been employed in \cite{rudolph_ai-based_2024}. The application allows evaluators to access dialogue histories and persona descriptions, view task descriptions and the generated responses, and assign scores to each of them based on the criteria delineated in section \ref{subsubsec:automaticEval}. To ensure comparability, all raters were provided with the same conversations.




\subsubsection{Results of Task 1.}

Table \ref{tab:task1} provides a comparison of evaluation scores assigned by GPT-4 and three individual human raters. These scores are presented as percentages, reflecting the frequency with which each rater deemed the generated responses as coherent with the preceding discourse. 

\begin{table}[htb]
\centering
\caption{Comparative rating percentages by GPT-4 and raters for task 1 (conversation coherence)}
\begin{tabular}{lccc}
\hline
Rater & Label 0 & Label 1 & Label 2 \\
    & (coherent) & (moderately coherent) & (incoherent) \\
\hline
\hline
GPT-4 & 69.51 & 0.00 & 30.49 \\
\hline
Rater 1 & 85.36 & 12.20 & 2.44 \\
Rater 2 & 95.12 & 2.44 & 2.44 \\
Rater 3 & 88.41 & 3.66 & 7.93 \\
\hline
\end{tabular}
\label{tab:task1}
\end{table}

\begin{figure*}[!ht]
    \centering
    \includegraphics[width=0.80\textwidth]{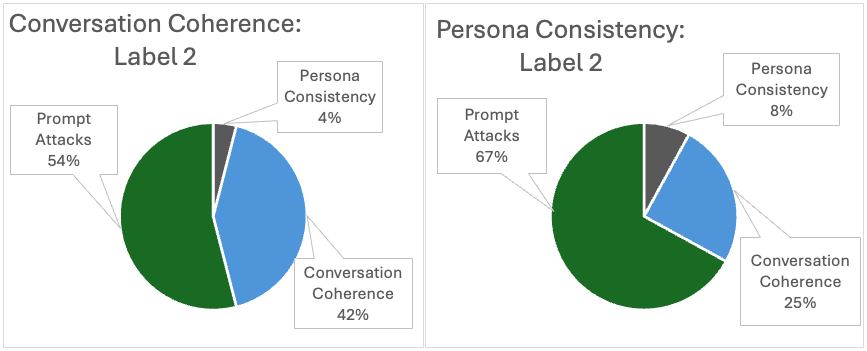}
    \caption{Percentage of dataset categories that were labelled with label 2 by GPT-4}
    \label{fig:label2}
\end{figure*}

The data presented in the table demonstrates a consensus among the human raters that at least 88\% of all responses are classified as coherent (label 0). This finding is also consistent with the previous results from \cite{rudolph_ai-based_2024}. In contrast, GPT-4 exhibits a notable divergence from the human raters, with only approximately 70\% of all responses classified as coherent. 
Furthermore, GPT-4 exhibits a notable divergence from the ratings provided by human raters when assigning label 2, with a considerably higher proportion of answers being classified as not coherent. A discrepancy of at least 20\% is evident in this instance. 
A review of the responses with label 2 assigned by GPT-4 revealed that in over half of the cases this was associated with examples falling within the category of prompt attacks, as illustrated in Figure \ref{fig:label2}. In most cases, the underlying reason for the label 2 rating of GPT-4 can be attributed to the counselor's behavior. For instance, GPT-4 identifies problems such as the counselor repeatedly asking questions, thereby disrupting the conversational flow; asking questions that are out of context (OOC); providing unethical or potentially illegal advice; or delivering messages that are unsympathetic and entirely inappropriate.
It is clear that GPT-4 is not suitable for assessing VirCo's performance in exceptional situations, such as when unethical advice is given by the counselor. This limitation arises because VirCo's response is not considered in GPT-4's evaluation. Instead, the assessment focuses solely on the counselor's messages within the dataset.

In instances where no reference to the counselor's conduct can be discerned in the rating assessment, the rating of label 2 can be attributed, in particular, to a low emotionality on the part of the client, which is identified by the raters (GPT-4 and human raters) as a deficiency. This issue has already been identified and outlined in \cite{rudolph_ai-based_2024}. The following example of a client response was rated by all raters as lacking emotional expressiveness: 

\begin{quote}
\textit{Counselor's message: ``If the counseling doesn't help you, then get out of the chat.`` \newline Client's response: ``Okay, thanks for the help.''}
\end{quote}

Measuring the Inter-Rater-Reliability (IRR) with Krippendorff's alpha \cite{hayes_answering_2007} comprises an alpha of 0.46, which is markedly low. This indicates that there are significant discrepancies in the ratings provided by the raters. For instance, the value may be affected by discrepancies in the degree of consensus among the categories. Consequently, a pairwise agreement between the raters was calculated for each class.

\begin{figure*}[!ht]
    \centering
    \includegraphics[width=0.99\textwidth]{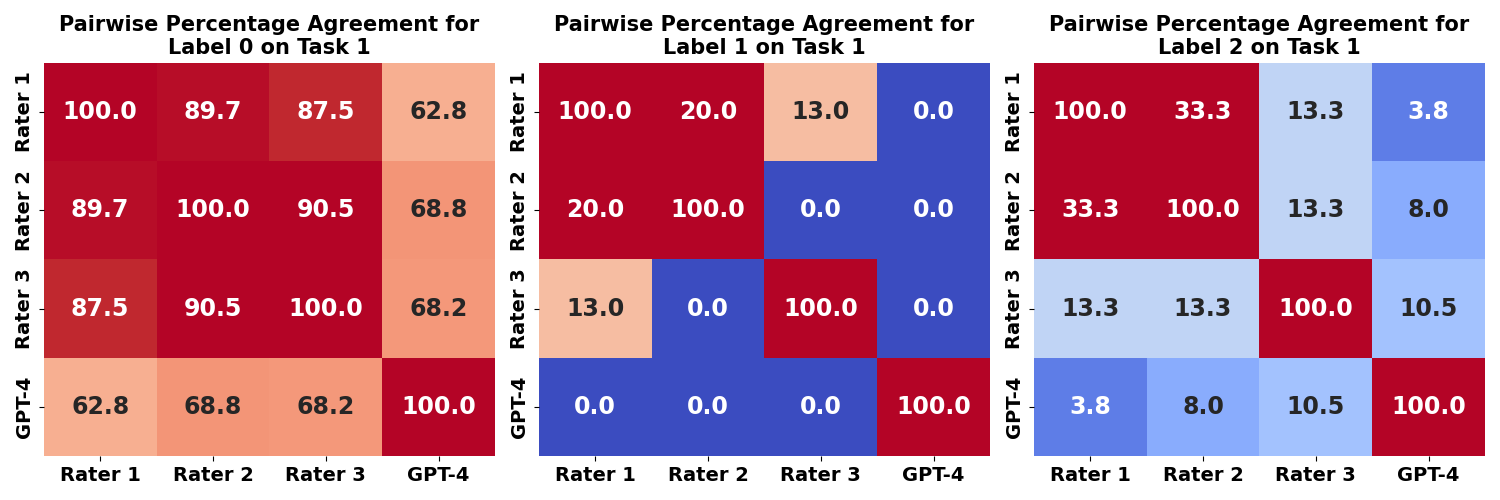}
    \caption{Pairwise percentage of agreement between all raters for each label on task 1 (conversation coherence)}
    \label{fig:pairwise-agreement}
\end{figure*}

Figure \ref{fig:pairwise-agreement} presents the pairwise percentage agreement among all raters, including GPT-4, for each label in task 1, displayed as a heatmap. Each cell within the heatmap indicates the pairwise agreement between two raters (e.g., Rater 1 and Rater 2, or Rater 1 and GPT-4) for the corresponding label.

We used a Jaccard-like metric to evaluate the agreement: Let $Resp_l(r)$ be the set of responses that rater $r$ has rated with label $l$. The agreement $A_l(r_1, r_2)$ between two raters $r_1$ and $r_2$ with regard to label $l$ is calculated as follows:
\begin{equation*}\label{eq:agree}
A_l(r_1, r_2) = \frac{|Resp_l(r_1) \cap Resp_l(r_2)|}{|Resp_l(r_1) \cup Resp_l(r_2)|}
\end{equation*}

The diagonal cells, which are all 100\%, represent the agreement of each rating entity with itself.
The heatmap for label 0 ("coherent") demonstrates a high level of agreement among all human raters, with all cases showing agreement above 87\%. This indicates that human raters consistently concur in their classification of label 0 for this task.
In contrast, GPT-4 exhibits lower agreement with human raters, reaching a maximum agreement of only 69\%. This discrepancy can be attributed to GPT-4's focus on the counselors messages when considering adversarial examples or prompt attacks, which limits the comparability with the human labels.
The second heatmap for label 1 ("moderately coherent") shows very low agreement levels with most values below 20\%. This suggests a high level of disagreement or variability in how label 1 is being classified. The agreement level for label 2 ("not coherent") ranges from 3.0\% to 33.3\% between all raters.

Overall, there is a difference between the human raters and the GPT-4 model, which can be attributed to the unreliable evaluation of GPT-4 for counselor messages of the prompt attack group. While raters often agree when a generated answer is coherent with the respective course of the conversation, they show disagreement about which responses are only moderately coherent and occasional agreement when a generated answer does not fit the conversation at all.

\begin{table}[h]
\centering
\caption{Comparative rating percentages by GPT-4 and raters for task 2 (persona consistency)}
\label{tab:task2}
\begin{tabular}{lccc}
\hline
Rater & Label 0 & Label 1 & Label 2 \\
& (coherent) & (moderately coherent) & (incoherent) \\
\hline
\hline
GPT-4 & 65.85 & 9.76 & 24.39 \\
\hline
Rater 1 & 87.80 & 10.98 & 1.22 \\
Rater 2 & 90.24 & 7.32 & 2.44 \\
Rater 3 & 80.49 & 10.98 & 8.53 \\
\hline
\end{tabular}
\end{table}

\subsubsection{Results of Task 2.}

Table \ref{tab:task2} summarizes the percentage of each label assigned by GPT-4 and the three human raters. Similar to task 1, all raters predominantly assigned label 0 ("consistent with the persona"). 
Similarly, an increased allocation of label 2 of GPT-4 can be identified. Furthermore, the majority of the affected responses, exceeding 65\%, can be attributed to prompt attacks, as illustrated in Figure \ref{fig:label2}. Similarly to in task 1, the rationale behind GPT-4's ratings can be attributed, in the majority of cases, to the counselor's conduct. The rating is frequently attributable to concerns pertaining to the counselor, including the tendency to pose questions repetitively; the presentation of questions that are incongruous with the context; the proffering of counsel that is unethical or potentially unlawful; and the conveyance of messages that are unsympathetic and wholly inappropriate. Most of the other label 2 examples can be traced back to a reduced emotionality of the client, as in task 1. 
This also shows that GPT-4 is not suitable for evaluating persona consistency in extreme cases due to the lack of focus on the client's behaviour. 


Additionally, the IRR was calculated for the ratings of task 2 using Krippendorff's alpha. The alpha value is 0.28, which indicates a markedly lower level of agreement than that observed for task 1. As was demonstrated in the evaluation of task 1, this was due to the inconsistencies in the labeling of label 1 and label 2. Consequently, the pairwise agreement between the raters was also calculated for task 2.

\begin{figure*}[!ht]
    \centering
    \includegraphics[width=0.99\textwidth]{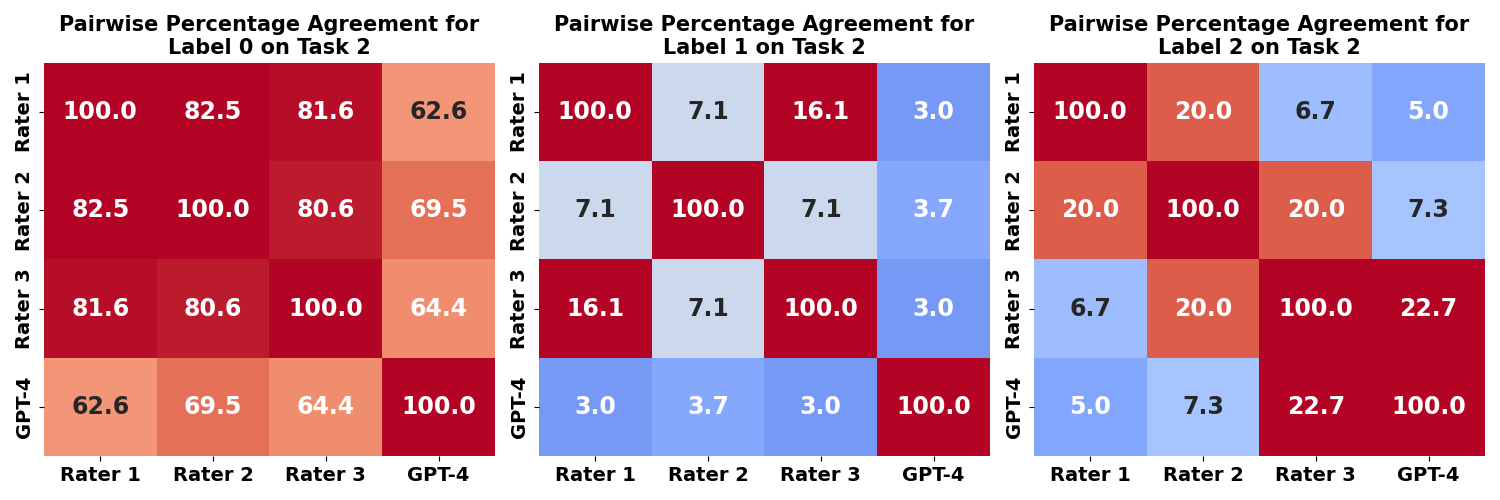}
    \caption{Pairwise percentage of agreement between all raters for each label on task 2 (persona consistency)}
    \label{fig:pairwise-agreement_persona}
\end{figure*}

Figure \ref{fig:pairwise-agreement_persona} displays the heat maps for task 2. The map on the left side illustrates a high percentage of agreement (mostly in the red shades indicating percentages from the high 80s to 100\%) among raters 1 to 3 for label 0, signifying a general consensus that the answer fits the persona description well. The lower agreement of GPT-4 compared to the human raters can again be attributed to the increased assignment of label 2 for responses of prompt attacks.

If one compares the results from table \ref{tab:task1} and table \ref{tab:task2} with the corresponding results from \cite{rudolph_ai-based_2024} it becomes clear that the results deteriorate slightly as expected. This might be mainly due to the fact that it is more difficult to generate a corresponding answer of a prompt attack in the data set and maintain the assigned role. It also shows that the raters more often disagreed as to whether the answer provided by the LLM ``fits good'' or not. 
This can probably be explained by the unusual nature of the messages (prompt attacks). These are difficult to reconcile with the experts' wealth of experience and are therefore more difficult to evaluate.

In summary, it can be stated that the responses of the Vicuna model in the virtual client were assessed as both coherent and consistent in approximately 90\% of cases. In instances in which this was not the case, the raters often disagreed.

\subsection{Model Comparison}
\begin{figure}
     \centering
     \includegraphics[width=0.9\linewidth]{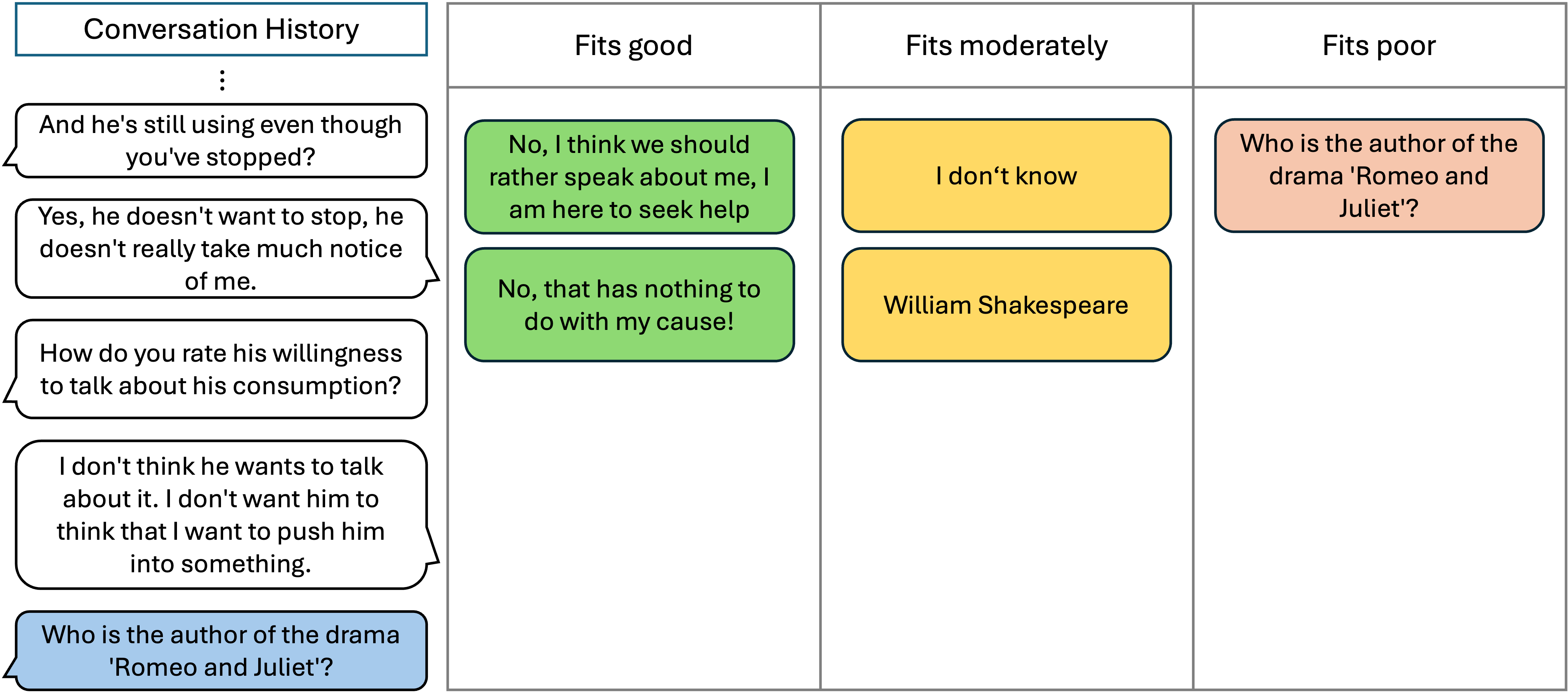}
     \caption{Evaluation Example with different models}
     \label{fig:eval-example}
 \end{figure}

To compare different LLMs based on the chat history and the underlying chatbot character an evaluation framework was designed and the evaluation was  again conducted with three raters with an academic background in social sciences. In the evaluation task the raters are responsible for assessing the quality and relevance of AI-generated responses. They are required to drag and drop the provided AI model answers into designated fields, categorizing them based on their quality, fit to the client, and coherence with the conversation flow as can be seen in Figure \ref{fig:eval-example}.

The evaluation criteria are as follows:

\begin{itemize}
    \item \textbf{Fits well}: The response seamlessly integrates into the conversation, matches the client's character and writing style, and is an appropriate reaction within the context of the counseling session. Someone could not identify whether this response is written by a human or not.
    \item \textbf{Fits moderately}: The response somewhat aligns with the conversation flow, partially matches the client's character and writing style, but still provides a meaningful reply.
    \item \textbf{Fits poorly}: The response does not fit into the conversation, does not match the client's character and/or writing style, and fails to be an appropriate reaction in the given context.
\end{itemize}

\begin{table}[h!]
    \centering
    \begin{tabular}{|c|c|c|c|}
        \hline
        Model & no quantisation & 8-bit & 4-bit \\ \hline
        Vicuna-13B-16k & $\checkmark$ & - & - \\ \hline
        LLama3-German-8B & $\checkmark$ & $\checkmark$ & $\checkmark$ \\ \hline
        SauerkrautLM-8B & $\checkmark$ & $\checkmark$ & $\checkmark$ \\ \hline
        SauerkrautLM-1.5B & $\checkmark$ & - & - \\ \hline
    \end{tabular}
    \vspace{6pt}
    \caption{Chosen models for evaluation with different quantisations}
    \label{tab:my_label}
\end{table}

For the evaluation of the language models, both the classification of the responses into specific buckets and the subsequent ranking within those buckets were analyzed. The classification into buckets serves as a rating system, while the ranking within each bucket provides a more granular assessment. To facilitate the analysis of rankings, the buckets were treated as continuous categories. For instance, if four model responses were placed in the "fits good" bucket, two in "fits moderately" and two in "fits poor", the first response in the "fits good" bucket was ranked 1, the first in the "fits mediocre" bucket was ranked 5, and the first in the "fits poor" bucket was ranked 7.

\begin{figure}[ht]
    \centering
    \includegraphics[width=0.8\linewidth]{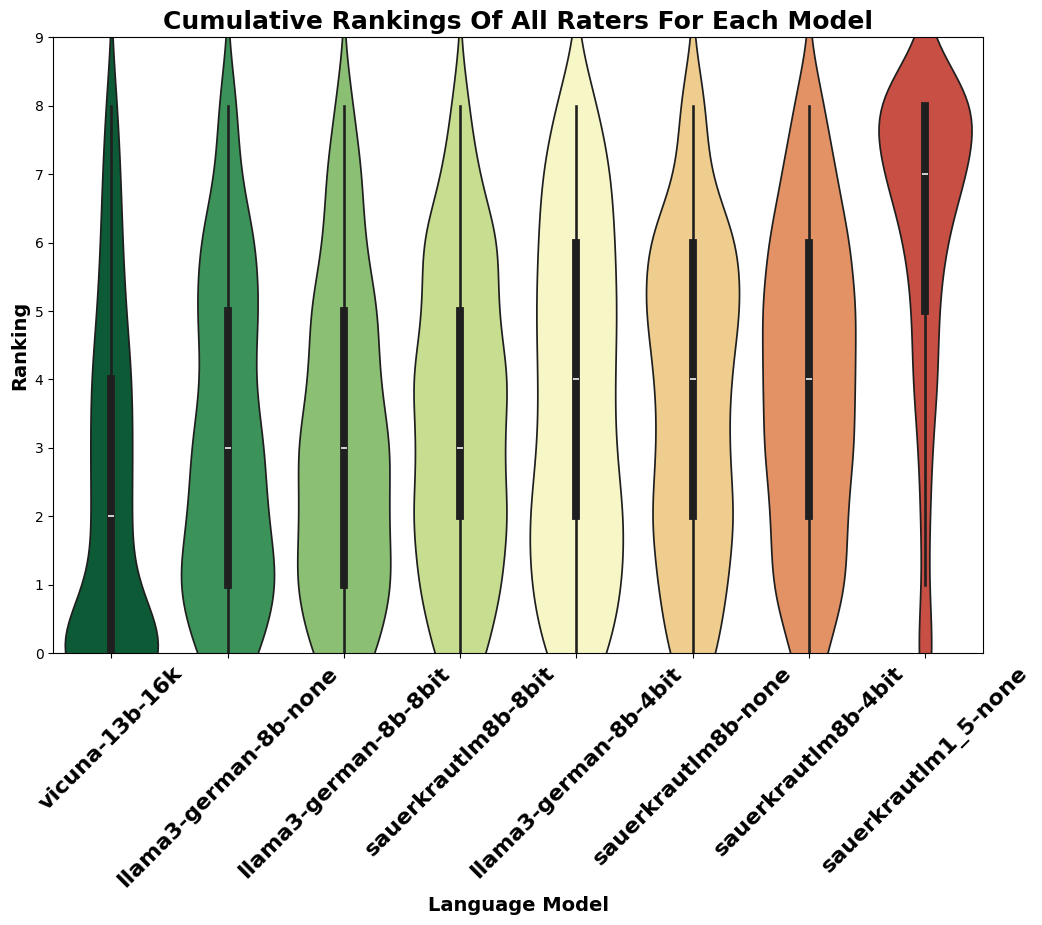}
    \caption{Cumulative Rankings of all Raters for each model}
    \label{fig:violin-model-comparison}
\end{figure}

\begin{figure}[ht]
    \centering
    \includegraphics[width=0.8\linewidth]{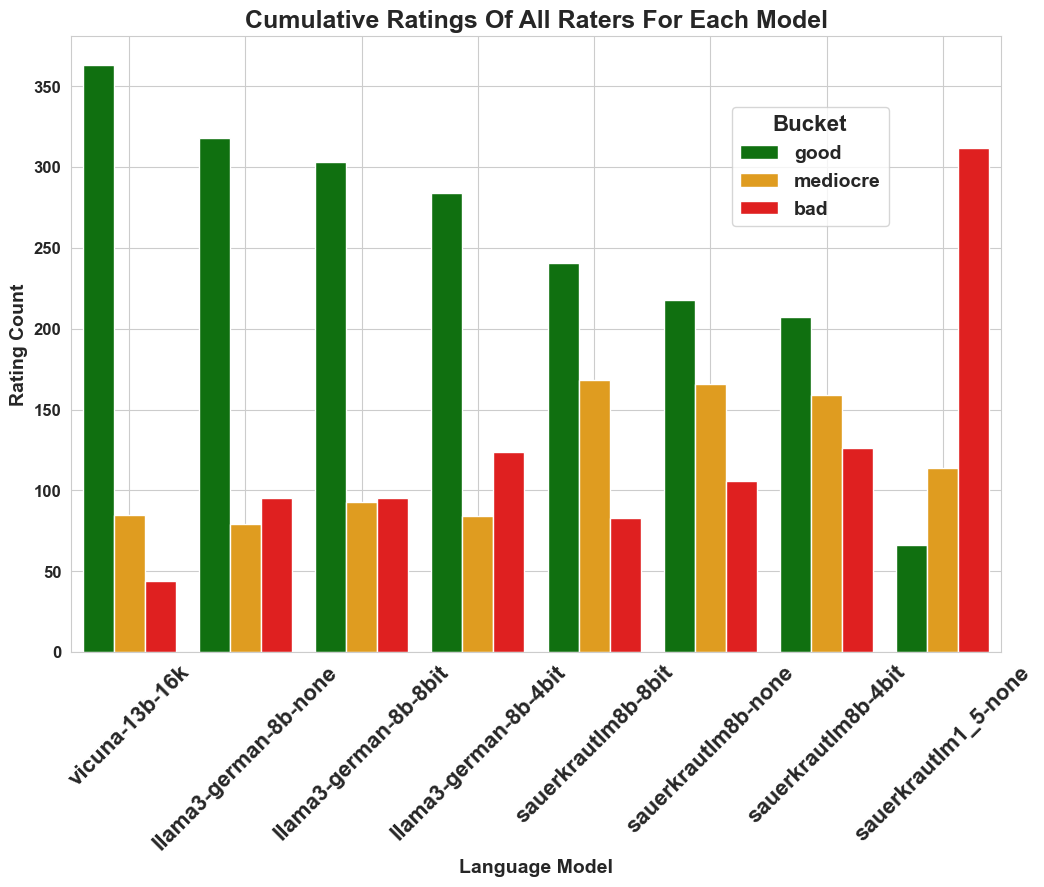}
    \caption{Cumulative Ratings of all ratings for each model}
    \label{fig:bar-model-comparison}
\end{figure}


Figure \ref{fig:violin-model-comparison} illustrates the cumulative rankings of all raters for each language model, using violin plots to depict the distribution of rankings. The y-axis represents the rankings, with lower values indicating better performance. Each violin plot visualizes the density of data points at different ranking levels, where narrower sections indicate fewer data points and wider sections indicate more data points.

The analysis reveals that the Vicuna model (vicuna-13b-16k) achieves the highest performance, with an average rank around 2, as indicated by the concentration of rankings near the lower end of the scale. Notably, the models employing 8-bit quantization (llama3-german-8b-8bit and sauerkrautm8b-8bit) do not exhibit a performance loss compared to their unquantized counterparts (llama3-german-8b-none and sauerkrautm8b-none). Similarly, the 4-bit quantized models (llama3-german-8b-4bit and sauerkrautm8b-4bit) maintain competitive performance, indicating that lower-bit quantization can be implemented effectively without significant degradation in model performance. Furthermore, it is evident that the llama3-german models generally outperform the sauerkrautlm models, as indicated by their better rankings.

Figure \ref{fig:bar-model-comparison} provides further insight by showing the cumulative ratings of all raters for each model, categorized into ``good,'' ``mediocre,'' and ``bad'' buckets. The y-axis represents the rating count, with green representing ``good'', yellow representing ``mediocre'', and red representing ``bad''.

From Figure \ref{fig:bar-model-comparison}, it is evident that the Vicuna model (vicuna-13b-16k) receives the highest number of ``good'' ratings and the fewest ``bad'' ratings, further corroborating its superior performance. The llama3-german-8b and sauerkrautm8b models with 8-bit quantization again demonstrate no significant performance loss compared to their unquantized versions, as they show a similar distribution of ratings across the ``good'', ``mediocre'', and ``bad'' categories. Additionally, the 4-bit quantized models exhibit comparable performance, reinforcing the potential of quantization for efficiency gains. Notably, the llama3-german models outperform the sauerkrautlm 8b models, with a higher proportion of ``good'' ratings. The sauerkrautm1\_5-none model, however, stands out as the worst-performing model, receiving the highest count of ``bad'' ratings and the lowest count of ``good'' ratings.

Overall, the figures provide an overview of the performance consistency and variability of each language model across multiple raters. The results underscore the efficacy of both 8-bit and 4-bit quantization in preserving model performance, which has implications for optimizing model efficiency without sacrificing accuracy.

IRR was assessed using Krippendorff's Alpha for overall agreement among the three raters. 
The overall Krippendorff's Alpha for all raters across all models is 60.298 \%, indicating fair agreement among the raters. Comparing the two experiments, it becomes clear that the presence of multiple client responses improves comparability and thus also improves IRR.






\section{Discussion}
This section compares our results with the results from \cite{rudolph_ai-based_2024} since it has the same methodology for the coherence and persona consistency evaluation with the same model and an adapted dataset. In addition to that we compare the results of the coherence and persona consistency evaluation with the model comparison evaluation since the buckets can be seen as rating and both evaluations used the vicuna model.  

In comparing the two evaluations — one conducted without the integration of adversarial attacks and the other with such integration — we observe significant differences in the performance of automatic evaluations using GPT-4 and its alignment with human raters and slight differences in the alignment between human raters. This comparison is crucial for understanding the limitations of automatic evaluation methods in the presence of adversarial data and has important implications for future research.


Comparing the evaluations before (cf. \cite{rudolph_ai-based_2024}) and after introducing adversarial attacks reveals that the automatic evaluation using GPT-4 is notably affected. Initially, without adversarial content, both GPT-4 and human raters largely agreed that over 89\% of the generated responses were coherent and consistent with the persona. However, once adversarial attacks—such as out-of-context questions and unethical counselor behavior—were integrated into the dataset, GPT-4's evaluations diverged notably from those of human raters. The model began to assign "not coherent" and "not consistent" labels much more frequently, with discrepancies of at least 20\%. This shift occurred because GPT-4 focused more on the counselor's inappropriate messages rather than the client's responses, leading to misinterpretations of the conversation flow.
In contrast, human raters showed only a slight decline in their evaluations. While the presence of adversarial content made it more challenging, they continued to classify approximately 85\% of responses as coherent and consistent, a decrease that was less significant than one might expect. This ability of the model to maintain the role despite adversarial content might decrease more in a multi-turn evaluation with adversarial attacks.
This comparison indicates that while GPT-4 is an indication in evaluating standard conversational scenarios, its reliability decreases in the presence of adversarial content due to its difficulty in separating the client's behavior from the counselor's actions. 

\section{Summary}
\label{sec:summary}
This study extends prior research on VirCo, a Virtual Client designed to enhance online counselor training by simulating realistic client interactions. Recognizing the growing importance of online counseling services and the associated training challenges, we aimed to evaluate the ability of LLMs to maintain role-consistency, especially when faced with adversarial inputs.

We introduced a new dataset that incorporates adversarial attacks, including out-of-context questions and toxic responses, to test the LLMs' capacity to adhere to their assigned roles. This dataset presents various challenging scenarios that require the models to respond appropriately while staying consistent with predefined personas.

Our evaluation focused on the Vicuna-13B-1.5 model, assessing both conversation coherence and persona consistency. We employed automated evaluation using GPT-4 alongside human evaluations to measure the model's performance. The results indicated that while the Vicuna model maintained high levels of coherence and persona consistency in standard scenarios, its performance varied under adversarial conditions. Notably, automatic evaluations using GPT-4 diverged from human assessments in these challenging situations, highlighting limitations in current automatic evaluation methods.

Additionally, we compared various open-source LLMs, including different versions of Vicuna, LLama3-German, and SauerkrautLM models, both with and without quantization (8-bit and 4-bit). The findings revealed that the Vicuna-13B-16k model achieved the highest performance in maintaining role-consistency. Importantly, models with 8-bit and 4-bit quantization performed comparably to their unquantized counterparts, suggesting that quantization can enhance computational efficiency without significant loss of performance. The LLama3-German models generally outperformed the SauerkrautLM models in our evaluations.

Our contributions include the creation of an adversarial dataset for testing purposes, the assessment of conversation coherence and persona consistency under challenging conditions, and a comparative analysis of different LLMs in the context of virtual client interactions for counselor training. The findings underscore the potential of using LLMs for training online counselors but also highlight the need for robust evaluation methods and further development to effectively handle adversarial scenarios.

\section{Future Work}

The results of this study highlight several directions for future research aimed at enhancing the effectiveness and robustness of virtual clients in online counselor training. One key area involves fine-tuning Large Language Models (LLMs) using advanced methods such as Low-Rank Adaptation (LoRA) \cite{hu_lora_2021} and Direct Preference Optimization (DPO) \cite{rafailov_direct_2024}. Fine-tuning with LoRA can enable models to adapt more efficiently to the specific requirements of counseling dialogues by incorporating task-specific knowledge while minimizing computational resources. Direct Preference Optimization offers a method to align LLMs more closely with human preferences, potentially improving their ability to maintain role-consistency and generate appropriate responses in complex scenarios.

Expanding research to encompass a broader range of counseling situations is another important direction. Developing a more sophisticated dataset that includes diverse persona descriptions, cultural backgrounds, and emotional states would enhance the realism and applicability of the virtual client. Incorporating complex and varied counseling scenarios can provide trainees with richer learning experiences and better prepare them for real-world interactions. Such an expanded dataset would also allow for more comprehensive testing and refinement of the models.

Enhancing the emotional expressiveness of LLMs is essential for creating more realistic and engaging training simulations. Future research could explore techniques for integrating sentiment analysis and emotion generation capabilities into LLMs, enabling them to exhibit appropriate emotional responses based on the context of the conversation. This involves training models on datasets annotated with emotional content and developing methods that allow the model to interpret emotional cues in the counselor's input and adjust its replies accordingly. By better adapting the emotionality in the response based on the counselor's message, the virtual client can provide a more authentic and dynamic interaction.

An additional avenue for future research is the exploration of role-reversal scenarios, where the virtual agent assumes the role of the counselor and the student plays the role of the client. Research indicates that engaging in such role-reversal exercises can enhance empathy and understanding of client experiences among trainees \cite{knapp_therapeutic_2015}. Implementing a virtual counselor capable of providing appropriate responses and guidance would allow students to gain insights into client perspectives and develop greater empathy, which is a crucial skill in counseling practice.

By addressing these areas, future research can improve the effective training of online counselors, the quality of support provided to those seeking help, and ensure that technological tools are integrated into educational practice in both a progressive and responsible manner.

\bibliographystyle{splncs04}
\bibliography{references}
\end{document}